\documentclass[conference]{IEEEtran}
\IEEEoverridecommandlockouts

\usepackage{cite}
\usepackage{amsmath,amssymb,amsfonts}
\usepackage{booktabs} 
\usepackage{graphicx}
\usepackage{textcomp}
\usepackage{xcolor}
\usepackage{algorithm}
\usepackage{algpseudocode}
\usepackage{float}
\usepackage[utf8]{inputenc}
\usepackage[T5]{fontenc}  
\usepackage[english]{babel}
\usepackage{multirow}

\def\BibTeX{{\rm B\kern-.05em{\sc i\kern-.025em b}\kern-.08em
    T\kern-.1667em\lower.7ex\hbox{E}\kern-.125emX}}
\begin{document}

\title{Bridging the Semantic Gap: Contrastive Rewards for
Multilingual Text-to-SQL with GRPO\\}

\author{\IEEEauthorblockN{Ashish Kattamuri}
\IEEEauthorblockA{\textit{Independent Researcher} \\
Denver, USA \\
ashishkattamuri@gmail.com}
\and
\IEEEauthorblockN{Ishita Prasad}
\IEEEauthorblockA{
\textit{University of Michigan}\\
Seattle, USA \\
Isprasad@umich.edu}
\and
\IEEEauthorblockN{Meetu Malhotra}
\IEEEauthorblockA{
\textit{Harrisburg University of Science and Technology}\\
Harrisburg, USA \\
Fmeetu@my.harrisburgu.edu}
\and
\IEEEauthorblockN{Arpita Vats}
\IEEEauthorblockA{
\textit{Linkedin}\\
California, USA \\
arpita.vats09@gmail.com}
\and
\IEEEauthorblockN{Rahul Raja}
\IEEEauthorblockA{
\textit{Linkedin}\\
California, USA \\
rahul.110392@gmail.com }
\and
\IEEEauthorblockN{Albert Lie}
\IEEEauthorblockA{
\textit{Forward Labs AI}\\
New York, USA \\
albert@forwardlabs.ai}
}

\IEEEoverridecommandlockouts
\IEEEpubid{\makebox[\columnwidth]{979-8-3315-8704-8/25/\$31.00~\copyright2025 IEEE \hfill} \hspace{\columnsep}\makebox[\columnwidth]{ }}

\maketitle

\IEEEpubidadjcol

\begin{abstract}
Current Text-to-SQL methods are evaluated and only focused on executable queries overlooking the semantic alignment challenge both in terms of the semantic meaning of the query and the execution results correctness. Even execution accuracy itself shows significant drops when moving from English to other languages, with an average decline of 6 percentage points across non-English languages. We address these challenges by presenting a new framework that combines Group Relative Policy Optimization (GRPO) within a multilingual contrastive reward signal to enhance both task efficiency and semantic accuracy in Text-to-SQL systems in cross-lingual scenarios. Our method teaches models to obtain better correspondence between SQL generation and user intent by combining a reward signal based on semantic similarity. On the seven language MultiSpider dataset, finetuning the Llama-3-3B model with GRPO improved the execution accuracy up to 87.4\% (+26 pp over zero-shot) and semantic accuracy up to 52.29\% (+32.86 pp). Adding our contrastive reward signal in the GRPO framework further improved the average semantic accuracy to 59.14\% (+6.85 pp, up to +10 pp for Vietnamese). Our experiments showcase that a smaller parameter efficient 3B llama model finetuned with our contrastive reward signal outperforms a much larger zero shot 8B llama model with uplift of 7.43pp on execution accuracy from 81.43\% on 8B model to 88.86\% on 3B model and cutting close on semantic accuracy with 59.14\% on 3B vs 68.57\% on 8B just with 3,000 reinforcement learning training examples. These results demonstrate how we can improve the performance of Text-to-SQL systems with contrastive rewards for directed semantic alignment without requiring huge amounts of datasets for training.
\end{abstract}


\section{Introduction}
With the current rise in adoptions and integrations of LLMs and improvements in Text-to-SQL systems, the multilingual aspect of these systems will be crucial as it will enable users across the world to query databases in their native language. While the Text-to-SQL systems enable users to access and run queries on the database without experience in coding SQL, the models' capabilities to support multiple languages are still limited. Current approaches suggest that the models' performance is better in English than other languages, with notable performance gaps observed in multilingual settings. The language models like Llama-3 show good results for multilingual generation but still struggle with cross-lingual consistency\cite{b1}.

Current approaches for improving multilingual Text-to-SQL systems are limited to reinforcement learning methods like GRPO\cite{b11} that focus primarily on mathematical reasoning tasks, while specialized prompting techniques like domain-specific knowledge injection\cite{b9} and chain-of-thought prompting\cite{b13} lack robust feedback mechanisms for semantic alignment. We address these limitations by proposing a novel method with a multilingual contrastive reward with Group Relative Policy Optimization (GRPO) to address this semantic alignment challenge and improve the execution and semantic accuracy of the Text-to-SQL systems. The new contrastive reward signal is used for training the Text-to-SQL model to score how closely the generated SQL query aligns with the intent of the users natural language query. We trained XLM-RoBERTa encoder\cite{b5} to create embeddings in the shared semantic space of the questions and the SQL queries for computing the contrastive reward for the generated SQL query. We ran our experiments on MultiSpider dataset\cite{b7} with 7 languages (Vietnamese, Spanish, Japanese, German, English, Chinese, and French) and achieved remarkable improvements with the integration of our contrastive reward signal in the GRPO framework giving us an increase in accuracy by 10 percentage points (pp) for Vietnamese and 6 pp on average on all other languages combined. In this paper we present our novel contrastive reward signal and the scalable architecture using XLM-RoBERTa encoder to compute the contrastive reward for the generated SQL query, and a lightweight training recipe that achieves 8B-level performance with a 3B model using only 3,000 examples, making high-quality multilingual Text-to-SQL more accessible and resource efficient.

\section{Related Work}
The Text-to-SQL task has long been a key area of research in natural language processing with an emphasis mainly on the English language. Efforts like Spider\cite{b15} have enabled significant progress in the accurate translation of natural language questions to working SQL queries. This success has led to models achieving high degrees of execution accuracy and syntactic validity in monolingual English settings.

Extending this progress to multilingual contexts, however, presents a new challenge. Early approaches to multilingual Text-to-SQL heavily relied upon translation-driven frameworks, where queries written in languages other than English were first translated into English before the parsing process. While this process might seem straightforward, it is extremely prone to semantic imprecision and misalignment during the process of translation, often generating errors or uneven results. Evaluation on multilingual datasets shows performance drops for non-English language scenarios compared to English, supporting the need for stronger multilingual approaches.

In an effort to respond to these challenges, several approaches have been proposed. Domain-specific knowledge injection techniques\cite{b9} and chain-of-thought prompting methods\cite{b13} have shown promise in improving Text-to-SQL performance. Such approaches use schema-aware prompts and provide few-shot examples created directly for multilingual datasets. Though these methods support cross-lingual generalization to an extent, they remain inherently bounded by their fixed properties. Without a mechanism that supports learning from model execution or correcting errors, prompt-reliant methods fail to dynamically adjust, often falling short of ensuring semantic accuracy even while attaining syntactic accuracy.

Our proposed method aims to address these limitations by employing reinforcement learning (RL) supplemented by semantic feedback. Rather than relying solely on manually designed prompts, our framework allows models to learn from explicit reward signals that include not only execution outcomes but also create a closer semantic mapping to user goals. This allows for creating an adaptive feedback loop that can steer the model towards producing more accurate and generally applicable SQL for multiple languages. To achieve this goal, we leverage the underlying principles of contrastive learning, which has seen widespread use in multilingual learning of representations as well as zero-shot cross-lingual transfer\cite{b3,b8}. In particular, we adapt methods from proven approaches like InfoXLM\cite{b3} and LaBSE\cite{b8} to build a multilingual contrastive encoder from XLM-RoBERTa\cite{b5}.

In contrast to past uses of reinforcement learning in semantic parsing using methods like REINFORCE\cite{b14} and PPO\cite{b10}, these methods often suffer from instability when used in combination with large language models. In an attempt to overcome this limitation, Group Relative Policy Optimization (GRPO)\cite{b11} has been proposed as an improved and stable option using KL-regularized updates to increase training stability and policy consistency. Originally developed for mathematical reasoning tasks, GRPO provides a more memory-efficient alternative to PPO while maintaining training stability. In this study, we build upon the principles of GRPO by combining it with our contrastive semantic reward, enabling the model to achieve optimal execution accuracy and semantic consistency in an efficient and stable manner.

In summary, while it is interesting to see that large foundation models like GPT-3\cite{b2}, PaLM\cite{b4}, and LLaMA\cite{b12} exhibit proficiency across a variety of tasks in zero-shot settings, their robustness can be limited in multilingual Text-to-SQL settings. Our study shows that a relatively smaller model, LLaMA-3B, when fine-tuned using semantically aware reinforcement learning, matches or outperforms larger zero-shot counterparts like LLaMA-8B. This highlights both the importance of semantic rewards and the effectiveness of our method in facilitating smaller models to attain a high level of accuracy in multilingual semantic parsing.

\section{Methodology}
\subsection{Approach Overview}

\begin{figure}[t]
    \centering
    \includegraphics[width=\columnwidth]{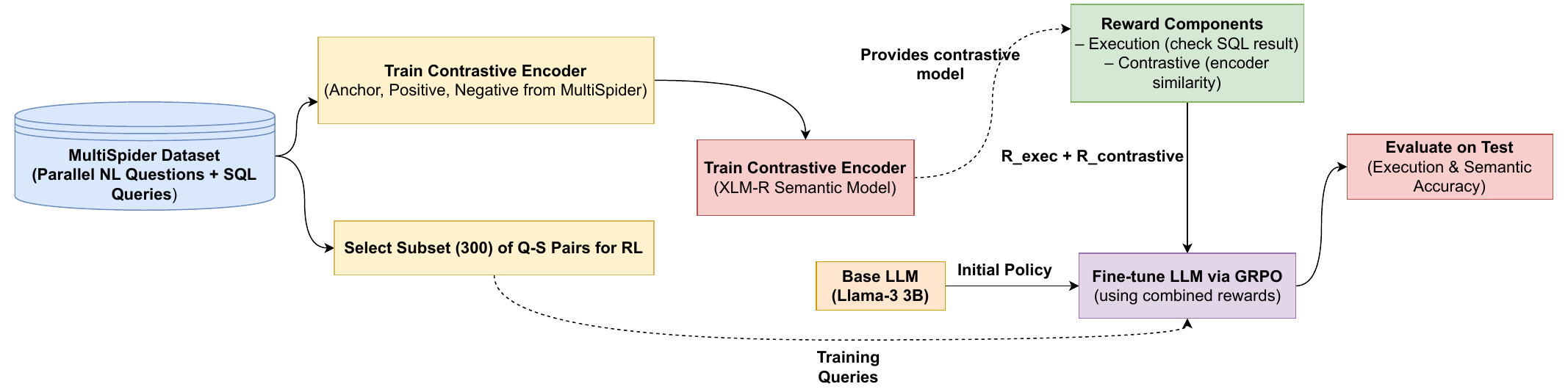}
    \caption{End-to-End Training and Evaluation Pipeline. The process begins with the MultiSpider dataset with parallel NL questions in 7 languages and corresponding SQL. A multilingual contrastive encoder is trained to align semantically equivalent questions across languages. This encoder provides a contrastive semantic reward signal during policy fine-tuning. The Generalized Reward Policy Optimization (GRPO) fine-tuning uses: (1) an execution reward ($R_{\text{exec}}$) giving a binary signal if the SQL yields correct results, and (2) a contrastive reward ($R_{\text{contrastive}}$) based on cosine similarity between question embeddings. These rewards update the LLM policy to produce SQL that is both executable and semantically faithful to the user's intent.}
    \label{fig:architecture}
\end{figure}

In our approach we fine-tuned a pre-trained Llama-3 3B model with reinforcement learning, using a novel contrastive reward that encourages consistent meaning across languages. 
Figure \ref{fig:architecture} illustrates our complete framework, including the novel contrastive reward mechanism that complements traditional execution and syntax rewards. 

The key insight of our method is that for multilingual Text-to-SQL, it's just not enough to check the execution completeness of the query, but we also need to ensure that the query is semantically aligned with the user question/input and the results actually match the gold SQL results. 

\subsection{Generalized Reward Policy Optimization (GRPO)}

We employ Group Relative Policy Optimization (GRPO)\cite{b11}, a model-free reinforcement-learning algorithm designed for language-model fine-tuning. Originally developed for mathematical reasoning tasks in the DeepSeekMath framework, GRPO follows a policy-gradient approach, but its update rule is crafted to be more stable and memory-efficient than traditional PPO on large language models. At each step the policy is moved only a small distance from a fixed reference policy (the frozen pre-trained model), and an adaptive KL-divergence penalty automatically scales to keep distribution shifts in check. GRPO also evaluates rewards directly on generated samples, so it avoids the noise introduced by training a separate reward model and can incorporate arbitrary reward functions. This flexibility lets us combine execution and multilingual semantic rewards within a single, stable optimization framework. The GRPO objective is designed to achieve maximum expected reward while maintaining proximity to the reference policy as shown below. 

\vspace{1em}
\begin{figure}[h!]
\centering
\fbox{
\begin{minipage}{0.85\columnwidth}
\begin{equation*}
\max_\theta \mathbb{E}_{x \sim \mathcal{D}}[\mathbb{E}_{y \sim \pi_\theta(y|x)}[R(x,y)]] - \beta D_{\text{KL}}(\pi_\theta||\pi_{\text{ref}})
\end{equation*}
\end{minipage}
}
\end{figure}
\vspace{1em}

where $x$ represents input (NL query + schema), $y$ is the generated SQL, $R$ is our combined reward function, and $\beta$ controls the KL penalty strength.

\subsection{Contrastive Reward Design}

Rather than simply using execution success or failure, we supply the model with a continuous semantic reward during training. We compute the cosine similarity between the meaning embedded into the input query and the meaning embedded into the gold-standard English query. This similarity (–1 to 1) is then used as a reward: the higher when the model's interpretation is closer to the actual intention.

Naturally, this reward doesn't explicitly rely on the SQL output; it's a directional cue that informs the model "Yes, you answered the question correctly" or "No, you may be off the mark". Coupling this with execution-based rewards, the model now receives feedback on both what to understand (question meaning) and what to do (act correctly).

\subsection{Contrastive Encoder Architecture and Training}

\subsubsection{Architecture}
Our reward encoder $E_C$ is built on XLM-RoBERTa-base \cite{b5}, a 12-layer transformer pre-trained on text from 100 languages. We enhance this with a 2-layer projection head (hidden dim 256 + ReLU) to produce 256-dimensional unit embeddings. Following the approach in BERT \cite{b6}, we use the CLS token's final representation as the sentence embedding, which is then passed through the projection head and L2-normalized.

\subsubsection{Training Data}
We use the MultiSpider dataset\cite{b7} to construct training triples. Two questions, an English question $Q^{(en)}$ and a Japanese question $Q^{(ja)}$, are regarded as a positive pair if they have the same SQL. For a negative, we use a question of another SQL (typically one that might lexically or structurally confuse the encoder).

\subsubsection{Loss Function}
We train $E_C$ with a triplet margin loss (margin 0.5), so that an anchor question and its translation (positive) are closer than any anchor and an unrelated question (negative) by at least the margin, as defined in Equation (1):

\begin{equation}
\mathcal{L}_{\text{triplet}} = \max(0, d(a, p) - d(a, n) + \text{margin})
\end{equation}

where $d$ is the cosine distance, $a$ is the anchor, $p$ is the positive example, and $n$ is the negative example.

\subsection{Reward Integration Mechanism}

At each training step the actor receives four feedback signals. The execution reward $R_{\text{exec}}$ is set to 1.0 when the generated query exactly matches the reference answer and 0 otherwise. The syntax reward $R_{\text{syntax}}$ is again 1.0 if the query can be executed without error. To encourage correct use of the database schema, we add a schema-matching reward $R_{\text{schema}}$, which grows when the query selects the right tables and columns. Finally, a semantic reward $R_{\text{sem}}$ is computed as the cosine similarity produced by our multilingual contrastive encoder.

\begin{figure}[ht]
\centering
\includegraphics[width=\columnwidth]{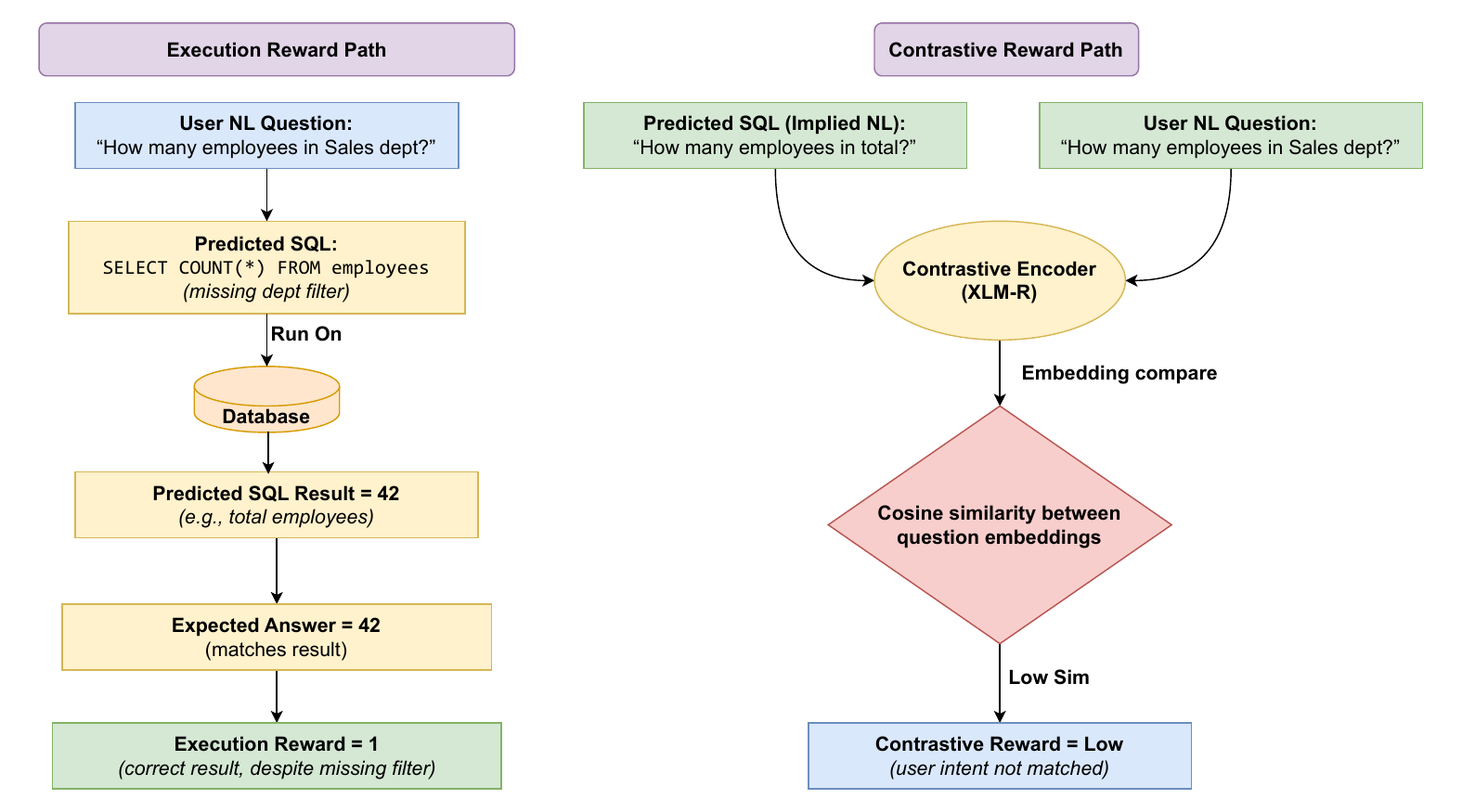}
\caption{Comparison of Execution Reward vs. Semantic (Contrastive) Reward. While execution rewards only verify if SQL queries run correctly on a database (binary signal), our contrastive rewards provide nuanced feedback about semantic alignment between user intent and generated SQL across languages, capturing subtle differences that execution signals miss. This is particularly important for multilingual queries where translation can introduce semantic ambiguities that binary execution signals cannot address.}
\label{fig:reward_comparison}
\end{figure}

Figure \ref{fig:reward_comparison} illustrates the key difference between traditional execution-based rewards and our contrastive semantic rewards. While execution rewards provide only a binary signal based on query correctness, contrastive rewards offer continuous feedback about semantic alignment, capturing subtle but important distinctions in meaning across languages.

In the current study we place full weight on $R_{\text{exec}}$ and treat $R_{\text{sem}}$ as a secondary signal that nudges the model toward better cross-lingual meaning preservation. An interesting direction for future work is to explore alternative weighting strategies—for example, increasing the influence of the semantic term—to learn whether a different balance among these rewards can further improve multilingual performance.

\subsection{Training Process}

\subsubsection{Dataset and Preprocessing}
We use MultiSpider \cite{b7}, which provides parallel queries across seven languages (Vietnamese, Spanish, Japanese, German, English, Chinese, and French). We use 3,000 training examples and evaluate on 50 examples per language from the dev set.

\subsubsection{Model Configuration}
We use the Llama-3 3B model with parameter-efficient fine-tuning via LoRA adapters (rank=16) targeting the attention layers. This approach significantly reduces memory requirements while maintaining performance.

\subsubsection{Training Loop}
Algorithm \ref{alg:training} outlines our complete training procedure. The model generates SQL candidates for each input, receives combined rewards, and updates its parameters via GRPO while maintaining KL constraints. The reward design integrates multiple complementary signals. The syntax reward ensures that generated SQL queries are well-formed by granting credit if they can be parsed and executed without errors. The schema-match reward measures whether the query correctly references the intended schema elements, rewarding valid table and column usage while penalizing irrelevant or nonexistent references. The execution reward captures semantic fidelity by comparing the execution results of the candidate query with those of the ground-truth query, assigning the highest score when the outputs match. Together, these rewards guide the model toward producing syntactically valid, schema-aware, and semantically faithful SQL.

\begin{algorithm}
\caption{Contrastive GRPO for Multilingual Text-to-SQL}
\label{alg:training}
\begin{algorithmic}[1]
\Require Base LLM $\pi_{\text{ref}}$, contrastive encoder $E_C$, dataset $\mathcal{D}$
\State Initialize policy $\pi_\theta \leftarrow \pi_{\text{ref}}$ with LoRA adapters
\For{iteration $= 1$ to max\_iterations}
    \State Sample batch $(Q_L, S, D) \sim \mathcal{D}$ (query, gold SQL, schema)
    \State Generate SQL candidates $\hat{S} \sim \pi_\theta(\cdot|Q_L, D)$
    \State Compute rewards:
    \State $\quad R_{\text{exec}} \leftarrow$ ExecutionReward($\hat{S}$, $D$)
    \State $\quad R_{\text{syntax}} \leftarrow$ SyntaxReward($\hat{S}$)
    \State $\quad R_{\text{schema}} \leftarrow$ SchemaMatchReward($\hat{S}$, $D$)
    \State $\quad R_{\text{sem}} \leftarrow$ CosineSimilarity($E_C(Q_L)$, $E_C(Q_{\text{ref}})$)
    \State $\quad R_{\text{total}} \leftarrow$ CombineRewards($R_{\text{exec}}$, $R_{\text{syntax}}$, $R_{\text{schema}}$, $R_{\text{sem}}$)
    \State Update $\pi_\theta$ using GRPO with $R_{\text{total}}$ and KL penalty to $\pi_{\text{ref}}$
\EndFor
\State \Return Fine-tuned policy $\pi_\theta$
\end{algorithmic}
\end{algorithm}

\subsubsection{Evaluation Metrics}
We assess performance using two complementary metrics:
\begin{itemize}
    \item \textbf{Execution Accuracy (ExecAcc)}: Percentage of generated SQL queries that execute without errors.
    \item \textbf{Semantic Accuracy (SemAcc)}: Percentage of generated SQL queries that are semantically equivalent to the gold standard, determined by comparing query results across multiple database states.
\end{itemize}

\subsubsection{Baselines and Models}
We compare the following configurations:
\begin{itemize}
    \item \textbf{L3B-ZS}: Zero-shot Llama-3 3B model
    \item \textbf{L8B-ZS}: Zero-shot Llama-3 8B model
    \item \textbf{L3B-GRPO-NC}: Llama-3 3B fine-tuned with GRPO (no contrastive reward)
    \item \textbf{L3B-GRPO-C}: Our full approach with contrastive reward
\end{itemize}

\subsubsection{Detailed Experimental Setup}
\label{sec:exp_details}
To ensure reproducibility of our results, we provide comprehensive details about our training infrastructure and hyperparameters:

\paragraph{Infrastructure:} Training was conducted on a single NVIDIA A100 GPU (40GB), with XLM-RoBERTa encoder training taking approximately 1 hour and GRPO fine-tuning taking 8-10 hours for 3000 training steps.

\paragraph{XLM-RoBERTa Encoder Training:} The contrastive encoder was trained for 2 epochs with a batch size of 96, learning rate of 2e-5, weight decay of 0.01, dropout of 0.1 in the projection head, and a warmup of 500 steps, using the AdamW optimizer. The triplet margin loss used a margin value of 0.5.

\paragraph{GRPO Fine-tuning:} For the policy fine-tuning phase, we used LoRA adaptation with rank=16 on the query, key, value matrices and output projection matrices of the self-attention layers. GRPO fine-tuning ran for 10 epochs over 3000 training steps, with a learning rate of 5e-6, batch size of 16, warmup of 500 steps, gradient clipping at 1.0, and a KL penalty coefficient $\beta = 0.02$. The reward weighting combined execution, syntax, schema, and semantic rewards with weights 1.0, 0.5, 0.5, and 0.2 respectively. These values were chosen heuristically to prioritize execution accuracy, while still encouraging syntactic validity, schema alignment, and semantic similarity. The work presented in this paper did not conduct an exhaustive hyperparameter search for optimal values; instead, the selected weights provided stable training across development runs. A more systematic study of adaptive or learned weight assignment is left for future work.

This methodology combines parameter efficient fine-tuning of a Llama-3 3B language model with our contrastive reward mechanism to improve cross-lingual semantic understanding. In the next section, we present the results of our experimental evaluation.

\section{Results and Discussion}
\begin{figure}[ht]
    \centering
    \includegraphics[width=\columnwidth]{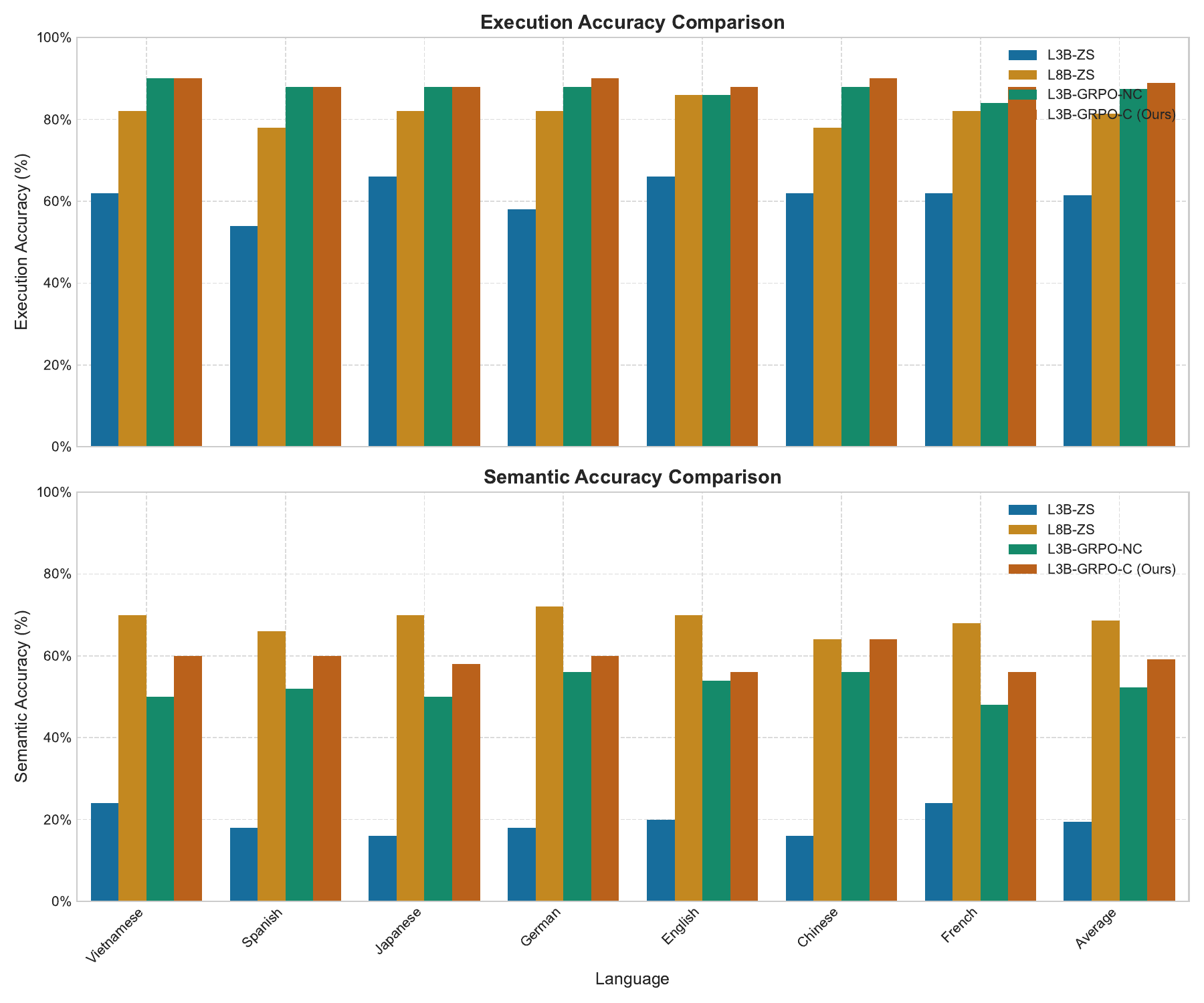}
    \caption{Execution Accuracy (top) and Semantic Accuracy (bottom) across different models and languages. Our L3B-GRPO-C model (dark blue) consistently outperforms the baseline L3B model and approaches or exceeds the larger L8B model, particularly in Execution Accuracy.}
    \label{fig:combined_results}
    \end{figure}

We evaluate our approach on the MultiSpider development set across seven languages. Table~\ref{tab:main_results} and Figure~\ref{fig:combined_results} present the core findings.

\begin{table*}[ht]
\centering
\caption{Text-to-SQL performance on MultiSpider (Dev Set). Scores are ExecAcc / SemAcc (\%). L3B-ZS = Llama-3B Zero-Shot; L8B-ZS = Llama-8B Zero-Shot; L3B-GRPO-NC = L3B + GRPO (No Contrastive); L3B-GRPO-C = L3B + GRPO (With Contrastive - Ours).}
\label{tab:main_results}
\resizebox{\textwidth}{!}{
\begin{tabular}{l|cc|cc|cc|cc}
\toprule
\textbf{Language} & \multicolumn{2}{c|}{\textbf{L3B-ZS}} & \multicolumn{2}{c|}{\textbf{L8B-ZS}} & \multicolumn{2}{c|}{\textbf{L3B-GRPO-NC}} & \multicolumn{2}{c}{\textbf{L3B-GRPO-C (Ours)}} \\
& Exec & Sem & Exec & Sem & Exec & Sem & Exec & Sem \\
\midrule
Vietnamese (vi) & 62.0 & 24.0 & 82.0 & 70.0 & 90.0 & 50.0 & 90.0 & 60.0 \\
Spanish (es)    & 54.0 & 18.0 & 78.0 & 66.0 & 88.0 & 52.0 & 88.0 & 60.0 \\
Japanese (ja)   & 66.0 & 16.0 & 82.0 & 70.0 & 88.0 & 50.0 & 88.0 & 58.0 \\
German (de)     & 58.0 & 18.0 & 82.0 & 72.0 & 88.0 & 56.0 & 90.0 & 60.0 \\
English (en)    & 66.0 & 20.0 & 86.0 & 70.0 & 86.0 & 54.0 & 88.0 & 56.0 \\
Chinese (zh)    & 62.0 & 16.0 & 78.0 & 64.0 & 88.0 & 56.0 & 90.0 & 64.0 \\
French (fr)     & 62.0 & 24.0 & 82.0 & 68.0 & 84.0 & 48.0 & 88.0 & 56.0 \\
\midrule
\textbf{Average} & \textbf{61.43} & \textbf{19.43} & \textbf{81.43} & \textbf{68.57} & \textbf{87.43} & \textbf{52.29} & \textbf{88.86} & \textbf{59.14} \\
\bottomrule
\end{tabular}
}
\end{table*}

\textbf{Baseline Performance:} 
The Llama-3 3B model (L3B-ZS) achieved an average ExecAcc of 61.43\% and SemAcc of 19.43\% in zero shot setting. The Llama-3 8B model (L8B-ZS) achieves an average ExecAcc of 81.43\% and SemAcc of 68.57\% with better performance over L3B-ZS as expected.

\textbf{Impact of GRPO Fine-tuning (No Contrastive Reward):} 
The performance of the finetuned Llama-3B (L3B-GRPO-NC) with vanilla GRPO fine-tuning achieved an average ExecAcc of 87.43\% and SemAcc of 52.29\%.

\textbf{Efficacy of Contrastive Rewards (Our Full Model):} 
Including the multilingual contrastive reward (L3B-GRPO-C) in the GRPO training loop achieved an average ExecAcc of 88.86\% and SemAcc of 59.14\% showcasing the efficacy of our approach. This results shows that contrastive reward is effective in improving semantic understanding of the model especially in the Vietnamese language where we see a +10pp improvement in semantic accuracy.

\textbf{Overall Improvement and Efficiency:} Our final model (L3B-GRPO-C) shows a total improvement of +27.43pp in ExecAcc and +39.71pp in SemAcc over the L3B-ZS baseline. This substantial gain is achieved using only 3000 GRPO training examples, demonstrating the sample efficiency of our approach.

\textbf{Comparison with Llama-8B Zero-Shot:} Remarkably, our fine-tuned Llama-3 3B model (L3B-GRPO-C) outperforms the zero-shot Llama-3 8B model in average ExecAcc (88.86\% vs. 81.43\%, a +7.43pp difference). While the L8B-ZS showed a better average SemAcc (68.57\% vs. 59.14\%), our L3B-GRPO-C model significantly closes this gap, reducing the L8B-ZS advantage to 9.43pp. This indicates that our targeted fine-tuning approach using contrastive rewards can enables smaller models to achieve highly competitive, and in some cases better performance compared to larger models in a zero-shot setting, offering a more resource-efficient path to high performance.

\subsection{Qualitative Analysis}
To illustrate how our contrastive reward mechanism corrects semantic errors that execution accuracy alone would miss, we present a representative example from our Vietnamese experiments shown in Figure 4:

\begin{figure}[H]
\centering
\framebox{
\begin{minipage}{0.9\columnwidth}
\footnotesize
\textbf{Example: Vietnamese Query Translation}

\hrulefill

\textbf{Original NLQ (Vietnamese):} \\
Hiển thị tên của tất cả các diễn viên đã tham gia vào nhiều hơn 3 bộ phim.

\hrulefill

\textbf{English Translation:} \\
Show the names of all actors who have participated in more than 3 films.

\hrulefill

\textbf{L3B-GRPO-NC (No Contrastive):} \\
\texttt{SELECT actor.name FROM actor JOIN casting ON}\\
\texttt{actor.id = casting.actorid GROUP BY actor.id}\\
\texttt{HAVING COUNT(*) >= 3;}

\hrulefill

\textbf{L3B-GRPO-C (With Contrastive):} \\
\texttt{SELECT actor.name FROM actor JOIN casting ON}\\
\texttt{actor.id = casting.actorid GROUP BY actor.id,}\\
\texttt{actor.name HAVING COUNT(DISTINCT casting.movieid) > 3;}

\hrulefill

\textbf{Gold SQL:} \\
\texttt{SELECT actor.name FROM actor JOIN casting ON}\\
\texttt{actor.id = casting.actorid GROUP BY actor.id,}\\
\texttt{actor.name HAVING COUNT(DISTINCT casting.movieid) > 3;}
\end{minipage}
}
\caption{Example query where contrastive reward improves semantic accuracy while maintaining execution accuracy. The model without contrastive rewards uses $\geq$ instead of $>$ as specified in the question and doesn't count distinct movies.}
\label{tab:example_query}
\end{figure}

In this example, both models generate SQL queries that execute successfully and happen to return the same result set on the test database state (passing execution accuracy). However, the query from the model without contrastive rewards (L3B-GRPO-NC) has two semantic issues: (1) it uses \texttt{>=} instead of \texttt{>} as specified in the question, and (2) it counts all casting entries rather than distinct movies, which could give incorrect results if an actor appears multiple times in the same film. The contrastive reward guides the model (L3B-GRPO-C) toward capturing these semantic nuances correctly, producing SQL that precisely matches the user's intent.
This example demonstrates how execution accuracy alone can be misleading and the role of contrastive reward in capturing subtle semantic nuances.

\section{Ablation Studies}
\label{sec:ablation}

\subsection{Impact of Contrastive Rewards}
Table~\ref{tab:ablation_contrastive} isolates the contribution of the contrastive reward component by comparing L3B-GRPO-C with L3B-GRPO-NC. The data clearly shows that the contrastive reward consistently improves Semantic Accuracy across all seven languages, with an average gain of +6.85pp. The most significant gains are seen in Vietnamese (+10pp), Spanish (+8pp), Japanese (+8pp), Chinese (+8pp), and French (+8pp), demonstrating the reward's broad utility in enhancing semantic fidelity.

\begin{table}[h!]
\centering
\caption{Ablation: Semantic Accuracy (\%) Improvement from Contrastive Rewards (L3B-GRPO-C vs. L3B-GRPO-NC).}
\label{tab:ablation_contrastive}
\resizebox{0.48\textwidth}{!}{
\begin{tabular}{l|ccc}
\toprule
\textbf{Language} & \textbf{L3B-GRPO-NC} & \textbf{L3B-GRPO-C} & \textbf{$\Delta$ SemAcc} \\
\midrule
French (fr)     & 48.0 & 56.0 & +8.0 \\
Vietnamese (vi) & 50.0 & 60.0 & +10.0 \\
Chinese (zh)    & 56.0 & 64.0 & +8.0 \\
Japanese (ja)   & 50.0 & 58.0 & +8.0 \\
English (en)    & 54.0 & 56.0 & +2.0 \\
Spanish (es)    & 52.0 & 60.0 & +8.0 \\
German (de)     & 56.0 & 60.0 & +4.0 \\
\midrule
\textbf{Average} & \textbf{52.29} & \textbf{59.14} & \textbf{+6.85} \\
\bottomrule
\end{tabular}
}
\end{table}

Figure 5 illustrates the language-specific improvements achieved by our contrastive reward mechanism, clearly showing the substantial semantic accuracy gains across all seven languages, with particularly notable improvements for Vietnamese and other non-English languages that typically challenge large language models.

\begin{figure}[ht]
\centering
\includegraphics[width=\columnwidth]{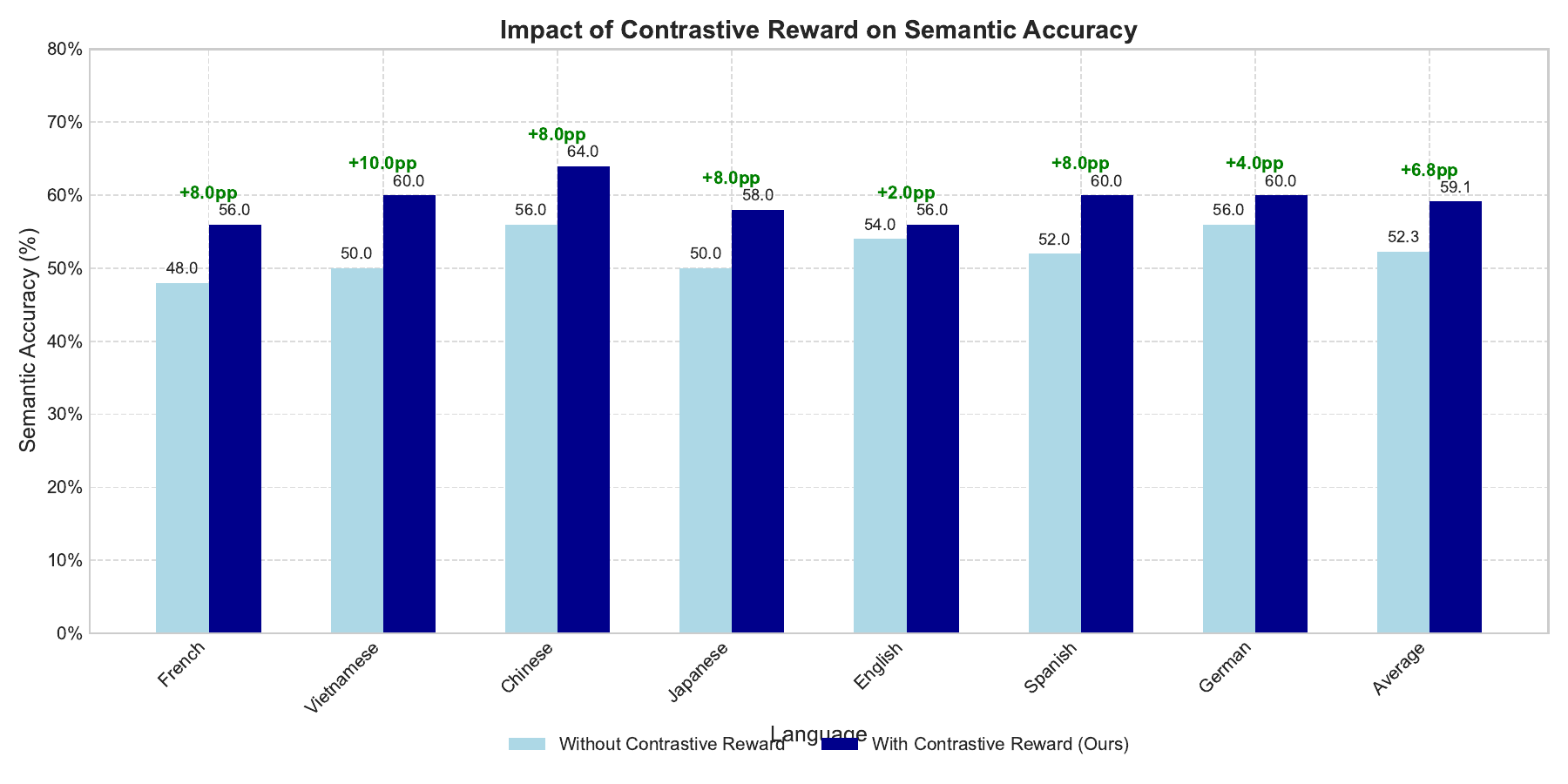}
\caption{Visual comparison of semantic accuracy with and without contrastive rewards across languages. Green annotations show the percentage point improvements achieved with our contrastive reward approach.}
\label{fig:ablation_chart}
\end{figure}

\subsection{Encoder Choice for Contrastive Reward}
\label{ssec:ablation_encoder}
To validate the choice of XLM-RoBERTa-base as our primary multilingual contrastive encoder, we conducted an ablation study where it was replaced by a smaller, less potent multilingual model, MiniLM-L6-v2 \cite{wang2020minilm}. The MiniLM model was trained using the same contrastive learning setup. 
Our initial experiments during encoder development showed that XLM-RoBERTa achieved significantly higher cross-lingual semantic similarity scores on held-out NLQ pairs compared to MiniLM. Specifically, XLM-RoBERTa attained approximately 0.9 average cosine similarity for positive pairs, while MiniLM only reached about 0.5.

When this MiniLM-based contrastive reward was integrated into the GRPO fine-tuning of Llama-3 3B, the resulting improvement in average Semantic Accuracy (over the L3B-GRPO-NC baseline) was only +1.5pp. This is substantially lower than the +6.85pp average SemAcc gain achieved when using the XLM-RoBERTa-based contrastive reward. This disparity underscores the importance of a strong multilingual encoder capable of capturing nuanced cross-lingual semantic equivalences to provide an effective reward signal. XLM-RoBERTa's extensive multilingual pretraining and larger capacity appear crucial for this task.

\section{Conclusion}
This research provided a novel concept for improving multilingual Text-to-SQL systems by merging a multilingual contrastive reward signal with Generalized Reward Policy Optimization (GRPO). Our method leverages a fine-tuned XLM-RoBERTa encoder to provide dense semantic similarity rewards, along with a Llama-3 3B model to build SQL queries that are both executable and semantically associated with user intent across seven different languages from the MultiSpider dataset. The efficiency of this method is shown by various experimentations. Substantial improvements above zero-shot performance are possible with standard GRPO fine-tuning alone, especially in execution accuracy. The model's semantic knowledge is further improved by integrating our contrastive reward system, with particularly evident improvements for languages that are typically difficult for large language models to recognize. Our method takes 3000 training samples to outperform larger models in prominent metrics. This effectiveness makes it possible to use equivalent approaches in even more languages and fields with limited comparable data. 

\section*{Acknowledgment}
In preparing this manuscript, we used grammar correction and rephrasing tools to improve clarity and readability of certain technical sections. All intellectual contributions, experimental design, results, and conclusions remain our original work.

\end{document}